\documentclass[conference]{IEEEtran}
\IEEEoverridecommandlockouts
\usepackage{cite}
\usepackage{amsmath,amssymb,amsfonts}
\usepackage{algorithmic}
\usepackage{graphicx}
\usepackage{textcomp}
\usepackage{xcolor}
\def\BibTeX{{\rm B\kern-.05em{\sc i\kern-.025em b}\kern-.08em
    T\kern-.1667em\lower.7ex\hbox{E}\kern-.125emX}}
\begin{document}

\title{Fake News and Hate Speech: Language in Common\\

\thanks{This work was partially supported by  European Commission under project IBERIFIER (CEF-TC-2020-2, European Digital Media Observatory) with reference 2020-EU-IA-0252 and partially supported by a Mar\'ia Zambrano grant of the Spanish Ministerio de Universidades and the European Union
NextGeneration\-EU/PRTR.\\ First author is also a member of \textit{Social Psychology Department} at the \textit{Universitat de València}}}

\author{\IEEEauthorblockN{Berta Chulvi}
\IEEEauthorblockA{\textit{PRLHT Research Center} \\
\textit{Universitat Politècnica de València}\\
Valencia, Spain \\
berta.chulvi@upv}
\and
\IEEEauthorblockN{Alejandro H. Toselli}
\IEEEauthorblockA{\textit{PRLHT Research Center} \\
\textit{Universitat Politècnica de València}\\
Valencia, Spain \\
ahector@prhlt.upv.es}
\and
\IEEEauthorblockN{Paolo Rosso}
\IEEEauthorblockA{\textit{PRLHT Research Center} \\
\textit{Universitat Politècnica de València}\\
Valencia, Spain \\
prosso@dsic.upv.es}
}

\maketitle

\begin{abstract}
In this paper we raise the research question of whether fake news and hate speech spreaders share common patterns in language. We compute a novel index, the ingroup vs outgroup index, in three different datasets and we show that both phenomena share an "us vs them" narrative.
\end{abstract}

\begin{IEEEkeywords}
Fake news, hate speech, NLP
\end{IEEEkeywords}

\section{Introduction}
Language use is central to the spread of fake news and hate speech. Both phenomena are part of information disorders according to the Council of Europe \cite{b1}. To stress the links between them some authors have used also the concept of harmful information \cite{b2}. Recently some studies have characterised the language used to disseminate fake news at the psycholinguistic level \cite{b3}. Other research has analyzed which psycholinguistic features are most present in the dissemination of online hate speech \cite{b4}. However, to the best of our knowledge, there is no empirical research showing that the two phenomena share linguistic patterns.

This research is based on the idea that hate speech and fake news are both at the service of online extremism. Conceptually, extremism involves hostility towards an apparent "foreign" group based on an opposing characteristic or ideology\cite{b5}. Following this line of reasoning, fake news and hate speech spreaders could share certain linguistic patterns if their deep goal was similar, that is to say, to divide the social arena between two groups (us vs them) focusing especially on the construction of otherness (them). 

It is well known that personal pronouns in language provide valuable information about subjects and their social environment \cite{b6}. It has been shown a higher occurrence of first and third-person plural pronouns in Extremist Alt-Right Subreddit forums \cite{b7}. Also, the use of third-person plural pronouns has been identified as a mark of extremism \cite{b8}. But all these research computed the frequency of first-person and third-person plural words independently, and none of them used a relational index to measure the relative emphasis placed on the ingroup or the outgroup. In Section III, we explain how we have created an index to measure this relative emphasis. 

\section{Datasets}
In the present research, we use three different datasets. 

\subsection{Check-worthiness tweets in CheckThat!2022}
The dataset was presented in CheckThat! lab in CLEF-2022 to determine whether a claim in a tweet is worth fact-checking. For Spanish, it was provided a corpus of 14,000+ posts from Twitter published by 310 politicians. This corpus was annotated by professional fact-checkers from Newtral\footnote{https://www.newtral.es}. They assigned the label 1 when considering that the tweet contains content that should be verified and 0 in the opposite case.  For the present research, we identify in the training and development part of this dataset (7,489 tweets) the authors of each tweet, and we select the tweets of 209 politicians that have tweets annotated in classes 0 (4,983 tweets) and 1 (2,184 tweets). See \cite{b9} for details about the corpus.

\subsection{Fake news spreaders in PAN2020}
The dataset was provided in the PAN Lab in CLEF-2020. The objective of this task was to determine whether or not the author of a Twitter feed is keen to spread fake news. In Spanish, a corpus with Twitter data was provided with 500 user feeds. For the present research, we use only the Spanish training dataset composed of 300 user feeds. We select the user feeds with have between 90 and 150 tweets, discarding users with an anomalous number of tweets. The result is a dataset that contains a total of 31,652 tweets from 284 user feeds: 146 users are considered fake news spreaders because they had shared at least once a piece of fake news and 138 users were labeled as non-fake news spreaders because they have not shared any fake news. See \cite{b10} for details about the corpus.

\subsection{Hate speech spreaders in PAN2021}
The dataset was created for PAN Lab in CLEF-2021 focusing on hate speech. The objective of the task was to determine whether or not the author of a Twitter feed is keen to spread hate speech, mainly against women and immigrants. In Spanish, a corpus of 300 user feeds (200 for training, and 100 for test) composed of the 200 last tweets of each user was provided. Users with more than ten hateful tweets were annotated as keen to spread hate speech and users that do not fit this condition as the opposite class. For the present research, we use only the Spanish training dataset composed of 200 users and a total of 40,000 tweets. Among them, 100 users are hate speech spreaders, and 100 are not. See \cite{b11} for details about the corpus.

\section {Methodology and hypothesis}
\subsection{Text representation}
Using the morphology information provided by the tagger spaCy\footnote{https://spacy.io} v3.0 we identified in the three datasets how many verbs and pronouns in first, second, and third person are contained in each tweet. We calculated the percentage of each person category with respect to the total of person tags present in each tweet. For each user, the scores of all tweets in each category were added and divided by the number of tweets of the user. In the CheckThat!2022 dataset we did this last operation twice considering two separate subsets of data for each user: (1) the tweets that should be reviewed and (2) the tweets that lack interest in being reviewed. As result, each Spanish politician included has two scores, one corresponding to the use of personal tags in tweets that should be checked and the other corresponding to the use of personal tags in tweets that are not interesting for being checked. Once we have this information for each user, we calculate the ingroup vs outgroup index as a subtraction between the use of the first person and the use of the third person in texts. If the score in this index is positive, it means that the user talks more about his group or about his own position (ingroup). Otherwise, if it is negative, it means that the user focuses their attention more on others (outgroup).

\subsection{Hypothesis}
In CheckThat!2022 our hypothesis is that tweets considered relevant for being checked will present a more negative score in the ingroup vs outgroup index than irrelevant tweets ($ H_{1} $). In PAN2020 dataset we expect that users with spread fake news present a more negative score in the ingroup vs outgroup index than users who do not ($ H_{2} $). The same is expected In PAN2021: the hypothesis is that users who spread hate speech present a more negative score in the ingroup vs outgroup index than users who do not ($ H_{3} $). 

\section{Results}
The scores of the ingroup vs outgroup index are not normally distributed according to the Kolmogorov-Smirnov test (p<.001 in the three datasets). Therefore, we performed a Wilcoxon signed-rank test for matched samples in the CheckThat!2022 dataset to test ($ H_{1} $) and a Mann-Whitney test in PAN2020 and PAN2021 datasets to test ($ H_{2} $) and ($ H_{3} $). Descriptive statistics are summarized in Table 1.

In the CheckThat!2022 dataset, the Wilcoxon signed-rank test for matched samples indicates that the ingroup vs outgroup index is statistically significantly lower in relevant tweets than in irrelevant tweets (Z=-8,995; p$<$.001). In PAN2020, the ingroup vs outgroup index is statistically significantly lower in users who spread fake news than in those who do not (U=7,648; p$<$.001). In PAN2021, the ingroup vs outgroup index is statistically significantly lower in users who spread hate speech than in those who do not (U=2,133; p$<$.001).

\begin{table}[!ht]
\caption{Ingroup vs outgroup index}
    \centering
    \begin{tabular}{|l|l|l|l|l|l|}
    \hline
        Data Collection & Indp. Var. & Users & Tweets & Mdn & Rank \\ \hline
        CheckThat! & irrelevant & 209 & 4,983 & -4.14 & 105.34 \\ 
        ~ & relevant & 209 & 2,184 & -16.45 & 99.23 \\ \hline
        PAN2020 & Fake news & 146 & 15,736 & -0.63 & 1.22 \\ 
        ~ & Not fake news & 138 & 15,916 & -0.51 & 0.96 \\ \hline
        PAN2021 & Hate  & 100 & 20,000 & -0.24 & 0.38 \\ 
        ~ & Not hate & 100 & 20,000 & -0.11 & 0.45 \\ \hline
    \end{tabular}
\end{table}

\section{Conclusion}
To the best of our knowledge, our study is the first that compares linguistic patterns used by fake news and hate speech spreaders. In our opinion, this research shows the usefulness of NLP for a deeper understanding of harmful information. As it has been shown, a vision of the social arena in terms of group conflict is underlying both phenomena. Empirical verification of these ``us vs them'' narratives could provide a mechanism to demand accountability from politicians and opinion leaders. For future work, we plan to check if it improves the explainability and performance of the classifiers used to detect automatically fake news and hate speech.

\end{document}